%% file: main.tex
\documentclass[runningheads]{llncs}
\usepackage[utf8]{inputenc}

\usepackage{cite}
\usepackage{amsmath,amssymb,amsfonts}
\usepackage{algorithmic}
\usepackage{graphicx}
\usepackage{textcomp}
\usepackage{xcolor}
\usepackage{ulem}
\usepackage{algorithm2e}
\usepackage{authblk}

\def\BibTeX{{\rm B\kern-.05em{\sc i\kern-.025em b}\kern-.08em
    T\kern-.1667em\lower.7ex\hbox{E}\kern-.125emX}}
\begin{document}

\title{Smart Camera Parking System With Auto Parking Spot Detection
}

\titlerunning{Smart Camera Parking System With Auto Parking Spot Detection}

\authorrunning{Tuan T. Nguyen et al.}
\date{August 2021}

\author{Tuan T. Nguyen$^\dagger$ \inst{1}
\and Mina Sartipi\inst{2}}

\institute{Center of Urban Informatics and Progress\\
  University of Tennessee at Chattanooga \\
  615 McCallie Ave, Chattanooga TN 37403\\ 
  \email{xwz778@mocs.utc.edu}\inst{1} 
\email{mina-sartipi@utc.edu}\inst{2}
 }
  



\maketitle
\input{0-abstract}
\input{1-Introduction}
\input{2-Related_Work}
\input{3-Methodology}

\input{4-Experiment_and_Result}
\input{5-Conclusion}

\input{6-Acknowledgement}

\bibliographystyle{splncs04}
\bibliography{main}

\end{document}

%% file: 0-abstract.tex
\begin{abstract}
Given the rising urban population and the consequential rise in traffic congestion, the implementation of smart parking systems has emerged as a critical matter of concern. Smart parking solutions use cameras, sensors, and algorithms like computer vision to find available parking spaces. This method improves parking place recognition, reduces traffic and pollution, and optimizes travel time.
In recent years, computer vision-based approaches have been widely used. However, most existing studies rely on manually labeled parking spots, which has implications for the cost and practicality of implementation.
To solve this problem, we propose a novel approach \textbf{PakLoc}, which automatically localize parking spots. Furthermore, we present the \textbf{PakSke} module, which automatically adjust the rotation and the size of detected bounding box. The efficacy of our proposed methodology on the PKLot dataset results in a significant reduction in human labor of 94.25\%. Another fundamental aspect of a smart parking system is its capacity to accurately determine and indicate the state of parking spots within a parking lot. The conventional approach involves employing classification techniques to forecast the condition of parking spots based on the bounding boxes derived from manually labeled grids. In this study, we provide a novel approach called \textbf{PakSta} for identifying the state of parking spots automatically. Our method utilizes object detector from PakLoc to simultaneously determine the occupancy status of all parking lots within a video frame. Our proposed method PakSta exhibits a competitive performance on the PKLot dataset when compared to other classification methods.

\end{abstract}

%% file: 1-Introduction.tex
\section{Introduction}
\label{sec:introduction}
According to the 2018 UN media report \cite{media}, 68\% of the world population will move to cities by 2050. The number of cars and other vehicles increases with urban population density, making parking management capacity and efficiency difficult. it increases air pollution and wastes drivers' time and energy. This causes parking lot vacancies and fluctuating occupancy rates. Thus, operators struggle to maximize facility revenue. 
Numerous contemporary studies \cite{zhong2023guided, sen2022bte, xu2023bits} primarily concentrate on traffic simulation as a means to alleviate traffic congestion. However, a recent INRIX survey \cite{inrix} found that the average American driver spends 17 hours per year searching for parking. In densely populated cities such as New York City, this number can reach 107 hours per year.  As the prevalence of autonomous vehicles rises, accurate and trustworthy information on parking lot layout and space availability is crucial. A Smart Parking System (SPS) is thus required for a sustainable urban environment. This platform connects drivers or autonomous vehicles with parking lot operators to everyone's benefit. A SPS should contain up-to-dated parking location data. A text message that suggests the best parking spaces to drivers not only reduces vehicle emissions, but also provides operators with consistent customers and increase revenue.

The main limitation on SPS functionality is the precision and efficacy of parking spot detection. Thus, sensors dominate the current SMS application. Because each sensor is designed for a single parking space, this strategy is too costly for future parking lot expansions despite its precision \cite{polycarpou2013smart}.
For example, the San Francisco's SF Park system program assists drivers in locating parking spaces in the city \cite{sfmta}. The program is funded to the tune of 27 million dollars, 19.8 million of which comes from the federal government. This substantial monetary contribution emphasizes the parking shortage in densely populated areas. Thanks to funding, SF Parking installed sensor technology in 19,250 parking spaces at an average cost of \$1,400 per sensor. The high cost comprises the price of the sensor in addition to the administration, labor, and logistical operations required to operate such a large system. 

Computer vision (CV) for detecting the status of parking spaces could be cost-effective. According to \cite{de2022systematic}, a single camera can cover multiple parking spaces, thereby eliminating the need for separate sensors. According to \cite{lin2017survey}, camera sensors are both cost-effective and unobtrusive. Additionally, the use of parking lot surveillance cameras has numerous advantages. They reduce installation and maintenance costs initially. In addition, these cameras detect improper parking, unusual behavior, and larceny, thereby enhancing parking management. In addressing such challenges, these capabilities outperform sensors. A variety of superficial to deep learning algorithms have been proposed for locating and categorizing parking spaces in this environment. \cite{biyik2021smart, de2022systematic} are systematic surveys.

As indicated in the survey paper \cite{de2022systematic}, computer vision is a highly promising methodology. However, it is observed that the majority of existing algorithms tend to approach the problem of parking lot detection as a binary classification task, focusing on the bounding boxes derived from manually labeled data. The parking spaces can be categorized as either occupied or vacant. In terms of performance, these techniques exhibit three primary limitations. First, the process of manually labelling parking spots is characterized by a significant expenditure of time and a lack of efficiency. Second, in the practical implementation of these solutions by a parking operator, it becomes necessary to re-annotate each parking spot to suit the new parking environment. This aspect significantly impacts the scalability of these systems. As an illustration, consider an operator responsible for overseeing five parking lots, each containing a minimum of 200 parking spaces. In order to utilize these solutions, it is necessary to carry out 1000 manual annotations. Furthermore, this process will need to be repeated whenever there is a change in the camera locations. Third, when the number of slots increases, the implementation of dependable deep-learning classification methods \cite{nyambal2017automated, valipour2016parking} necessitates the execution of numerous forward passes, resulting in sluggish real-time input to drivers and limited capacity for new activities. 

\begin{figure}[ht!]
\begin{center}
	\includegraphics[width=0.9\linewidth]{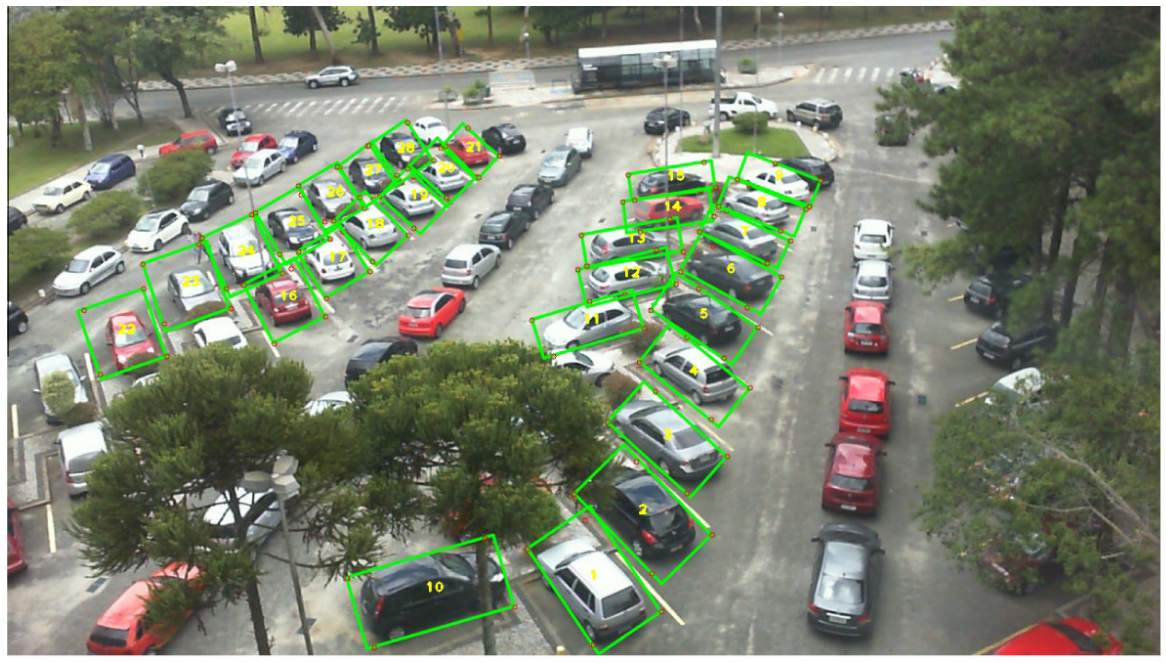}
	\caption{Parking Spots Detection - image from Pklot Dataset \cite{de2015pklot}}
\end{center}
\label{fig:parking_spot}
\end{figure}

In order to address the initial two challenges, we present a proposed automated parking space localization technique known as \textbf{PakLoc}. This method employs an object detection algorithm to accurately identify and locate parking spots. Our assumption is that the parking spots in the camera footage are likely to be the locations where cars are observed most frequently. It is demonstrated in figure \ref{fig:parking_spot}, where the green coordinates (defining parking spots) should be automatically defined.
Furthermore, we have introduced a skewness adjustment module, named \textbf{PakSke}, to address the issue of deep learning detectors generating bounding boxes that are perpendicular to the image borders. This problem arises due to the fact that many parking spots have different angles, as depicted in Figure \ref{fig:parking_spot}. The purpose of the PakSke module is to facilitate the automatic rotation of bounding boxes, thereby ensuring their alignment with the optimal angle of the corresponding labels. 
By conducting extensive experiments on PKLot datasets, we present empirical evidence that supports the effectiveness of the proposed method in detecting parking spaces without the need for prior knowledge. This approach eliminates the need for time-consuming manual labelling, hence streamlining the process. The results show that the method achieved an average recall (AR) at an IoU threshold of $0.75$ (AR75) up to 94.25\%.

In order to address the third problem, we propose the implementation of the \textbf{PakSta} framework, which utilizes an object detector in PakLoc to accurately determine the state of all parking spots concurrently. Our suggested system undergoes thorough testing using the widely recognized parking datasets PKLot, in order to conduct a comprehensive benchmark analysis. The findings obtained from our experiment demonstrate a superior level of competitive performance when compared to exhaustive classification approaches.

The primary contributions of this work are:
\begin{itemize}
    \item We provide the PakLoc algorithm as a means of autonomously identifying parking spaces. By utilizing camera video frames as input, PakLoc is able to considerably minimize 94.25\% (AR75) of the manual labeling effort required during the deployment of a SPS in a new parking lot environment.
    \item We propose the PakSke module, designed to streamline the process of automating the rotation of result bounding boxes. This procedure guarantees that the bounding boxes are aligned with the optimal angle of the relevant labels. Empirical study show that PakSke have postive impact on both PakLoc and PakSta. The potential exists for its use as a plug-in module within a comparable framework.
    \item We propose the PakSta framework, which aims to concurrently detect and monitor the state of all parking lots. PakSta employs the detector utilized in PakLoc, obviating the necessity for training an additional model.
\end{itemize}
The rest of the paper is structured as follows: related work is explained in Section \ref{sec:related_work}; Section \ref{sec:Methodology} presents our proposed methodology; Section \ref{sec:experiment} describes our datasets, evaluation metrics, experimental setups, and results; and Section \ref{sec:conclusion} presents a discussion and concluding remarks.

%% file: 2-Related_Work.tex
\section{Related Work}
\label{sec:related_work}
As discussed in Section \ref{sec:introduction}, we divided our framework into two sub-problems: 
\begin{itemize}
    \item \textbf{Automatic Parking Spots Localization} which tries to localize the parking spots automatically from parking lot video frames.
    \item \textbf{Parking Spots Status Identification} which predicts the status of each parking spot as either vacant or occupied.
\end{itemize}
In this section we will discuss the related works that solve the two problems mentioned above. 
\subsection{Automatic Parking Spots Localization}
\label{sec:Parking_Spots_Localization}
The determination of parking spot locations occurs when a new system is installed or when there is a change in the camera's perspective. The position of a parking spot is determined by four coordinates within a picture of the parking lot. Considerable efforts have been dedicated to addressing the issue of parking spot localization. These works can be categorized into three distinct categories: traditional image processing approaches \cite{bohush2018extraction, zhang2019smart, zhang2020image}, chess-board approaches \cite{nieto2018automatic, vitek2017distributed}, and deep learning approaches \cite{zhang2019smart, bohush2018extraction, nieto2018automatic, vitek2017distributed, agrawal2020multi, coleiro2020car, agrawal2020multi, coleiro2020car, hurst2020robust, kirtibhai2020faster, patel2020car, pannerselvam2021adaptive, wang2022global}.

For \textbf{traditional image processing approaches}, in the study \cite{zhang2019smart} a perspective transformation was employed, alongside the utilization of standard image processing techniques such as Canny and Gaussian edge detectors, in order to find parking spots. Similarly, in the work of \cite{bohush2018extraction}, an application of perspective transformation is employed in classical image processing techniques, with the objective of transforming the parking places into rectangular shapes that are parallel to the axes. The painted lines in parking lots were recognized using the utilization of Otsu's binarization technique and morphological processes. In another work \cite{zhang2020image}, using Canny edge detector and Hough transform, the authors employ conventional image processing techniques to identify the lines delineating each parking space. However, images of a vacant parking area with clearly discernible markings were used exclusively

For \textbf{chess-board approaches}, it is assumed that the parking spaces conform to a rectangular shape \cite{nieto2018automatic}. Subsequently, a homography matrix transformation is employed in conjunction with a standard video frame captured by a parking lot camera to compute the corresponding bird's-eye view representation of the parking lot. Ultimately, by considering the arrangement of rows and columns, as well as the coordinates of the corners of the rectangular shape, it becomes possible to automatically establish the boundaries of parking spaces. In another work \cite{vitek2017distributed}, The authors employed a chess-board strategy to extract features from Histogram of Oriented Gradients (HOG) in order to distinguish between automobiles and backgrounds. This was achieved by utilizing a classification algorithm. During the merging step, the process involves combining all chess-board squares that have been classified as cars in order to provide a unified map of parking spots. Nevertheless, a limitation of this study is that the authors rely on car detection to anticipate available parking places, which may result in misleading positive predictions if a car is merely passing by the parking lot. In our proposed methodology, we address the aforementioned issue by effectively monitoring the trajectory of an identified vehicle across a limited number of sequential frames. Specifically, we observe that a stationary vehicle exhibits a consistent intersection over union (IOU) value of 100\% across successive frames, but a moving vehicle demonstrates a reduced IOU value.   

For \textbf{deep learning approaches}, the \textbf{CNN based methods} are the primary trend of many studies. In \cite{agrawal2020multi, coleiro2020car}, The authors employ the Mask R-CNN architecture \cite{he2017mask} for the purpose of car detection and parking place localization.  Specifically, in \cite{agrawal2020multi}, the authors employed the identification of observed vehicles to extract parking spaces, making a notable assumption that regions where vehicles remain stationary for extended durations can be considered potential parking spaces. In a similar vein, the researchers in \cite{coleiro2020car} employed automobile bounding boxes to build a heat map for the purpose of localizing parking places. In another work \cite{hurst2020robust}, Satellite photography is employed for the purpose of extracting parking blocks. A parking block refers to a cohesive arrangement of parking spaces that can be identified in satellite imagery captured from a perpendicular perspective to the parking lot. These photos exhibit clear markings that delineate the boundaries of each individual parking space. In this particular circumstance, it is imperative that the parameters of the parking lot cameras align with the satellite photos obtained by surveillance cameras. The utilization of a U-Net network is employed for the purpose of detecting parking places. In recent studies \cite{kirtibhai2020faster, patel2020car},The authors employ the Faster R-CNN or YOLOv4 models for car detection. To ascertain the stationary status of a vehicle, the bounding boxes observed in two or three successive frames are compared. A parking place is defined as the location where a car is parked. The proposed approach from CNRPark-EXT was assessed by the authors across various weather conditions during a three-day period characterized by high activity levels. In another work \cite{de2023vehicle}, The authors deploy the Cascade Mask R-CNN method proposed by \cite{he2017mask} to carry out car segmentation in their study. Subsequently, they generate a heat map by aggregating the segmentation outcomes obtained from all video frames. The parking slots have been accurately identified and delineated using an overhead heat map with a well defined threshold as a parameter.

In addition to the widely used CNN model employed in deep learning approaches, it is important to recognize the advent of \textbf{vision transformers} as a different approach to image-related problems, as proposed by \cite{dosovitskiy2020image}. The input processing in visual transformers adheres to the methods outlined in the standard transformer encoder introduced by \cite{vaswani2017attention}. The representations are generated by transformers, which estimate the associations among image portions that are arranged in a linear sequence. A number of recent studies have employed the vision transformer model in order to tackle the problem of autonomous parking lot localization. In the work \cite{pannerselvam2021adaptive}, The authors present a novel transformer-based detection model that demonstrates the capability to recognize automobiles from various perspectives. Additionally, they enhance the car detector by introducing a module designed to improve illumination conditions specifically for low-light photos. In other study \cite{wang2022global}, The researchers introduced a novel module named Global Perceptual Feature Extractor (GPFE) within a transformer framework, aiming to attain global attention for a convolutional neural network (CNN) classification task. Recently, \cite{shi2022research} proposed a methodology that utilizes a Swing Transformer for the purpose of semantic segmentation in the context of parking spot localization. Furthermore, the Canny algorithm is employed to acquire the vehicle mold and improve the precision of detection. The researchers conducted computations in order to evaluate the association between the vehicle's mold and the manually drawn warning line. The aforementioned association is thereafter evaluated against a pre-established threshold value to determine whether the vehicle is situated inside the assigned area. 

Deep learning-based approaches commonly employ rectangular bounding boxes that are perpendicular to the borders of the image in order to detect objects. However, as depicted in Figure \ref{fig:parking_spot}, numerous parking spots exhibit varying angles in relation to the image border. All of the aforementioned methods fail to address this issue. In order to solve this problem, we propose the implementation of a skewness adjustment module referred to as \textbf{PakSke}. This module is designed to automatically rotate bounding boxes in order to align them with the optimal angle of the corresponding labels. Our backbone for detection model is deformable DETR \cite{zhu2020deformable} which exhibits the ability to detect objects across multiple scales.

\subsection{Parking Spots Status Identification}
The parking spot status identification task aims to predict the status of parking spots as occupied or vacant. The related works can be divided into two catagories: classification approaches and detection approaches. 

\subsubsection{2.2.1 Classification Approaches}
Most of the existing work uses the \textbf{classification approaches}, where the  parking spot status identification is treated as a classification task and each pre-defined parking spot is classified as occupied or vacant. The parking spots are defined manually or by a method mentioned in Section \ref{sec:Parking_Spots_Localization}. Then, the classification problem can be solved by feature extraction classification methods or a deep learning classification method.

In the feature extraction classification methods, the input image is pre-processed and extracts one or more feature vectors. The feature vector is fed into a traditional classification model such as Support Vector Machine (SVM), k-nearest neighbors (k-NN) or  Multilayer Perceptron (MLP). In the work \cite{almeida2013parking}, the authors use Local Phase Quantization (LPQ) \cite{ojansivu2008blur} and the Local Binary Patterns (LBP) \cite{ojala1999unsupervised} feature vectors and SVMs as classifiers to predict the status of parking spots. In other work \cite{suwignyo2018parking}, The Quadratic Local Regression Binary Pattern (QLRBP) \cite{lan2015quaternionic} method was utilized to extract texture features from the color images of the parking spaces. The authors employ k-nearest neighbors (k-NN) and support vector machines (SVMs) as classification algorithms. A total of 6000 parking spaces from the UFPR04 subset were utilized for the tests. In \cite{dizon2017development},  LBP and Histogram of Oriented Gradients (HOG) \cite{dalal2005histograms} were employed as feature descriptors in the context of a linear Support Vector Machine (SVM) classifier. The authors also utilized a background subtraction methodology incorporating the Adaptive Median Filter (AMF). The findings presented in this study demonstrate that a classifier trained solely using the Histogram of Oriented Gradients (HOG) method yielded favorable outcomes when applied to the UFPR04 subset from the PKLot dataset. Other works utilize the feature of pixel values under different color spaces, for example in the work \cite{baroffio2015visual}, the histograms in the Hue, Saturation, and Value (HSV) color space are computed directly with smart cameras. The histograms are transmitted to a central location, where they are utilized as features for a Support Vector Machine (SVM) classifier. The authors in \cite{goez2018automatic} offer a methodology that utilizes a bag of features to classify individual parking spaces. The researchers employ the Scale-Invariant Feature Transform (SIFT) \cite{lowe1999object} for feature extraction. Additionally, they utilize a Support Vector Machine (SVM) with a radial basis kernel as the classification model.

In the deep learning classification methods, the workflow has resemblance to that of feature-based classification methods, albeit with the integration of the feature extraction and classifier training components within a representation learning block. Transfer learning is a common technique, in \cite{nyambal2017automated}, LeNet \cite{lecun1998gradient} and AlexNet \cite{krizhevsky2012imagenet} are first pre-trained in a generic dataset like ImageNet and then fine-tuned in a parking lot dataset. Several studies \cite{acharya2018real, amato2018wireless, kolhar2021multi, merzoug2019smart, rahman2020convolutional} have put forth lightweight models that draw inspiration from established convolutional networks, including LeNet, AlexNet, and VGGNet \cite{simonyan2014very}. These custom models largely consist of convolutional networks that bear resemblance to the original networks, however with a reduced number of layers. Typically, these types are designed specifically for devices with limited processing capability and low-power requirements, such as smart cameras. The perspective transformation method, as described in previous works \cite{bohush2018extraction, bura2018edge, nieto2018automatic}, is employed to convert the representation of the parking lot into a two-dimensional grid format. Nevertheless, due to the strong reliance of the perspective projection process on the camera configuration in the parking lot, it becomes necessary to retrain classification models for various camera configurations. This raises concerns over the scalability of such systems. In order to address this issue, the authors in \cite{li2017uav} have proposed a Generative Adversarial Network (GAN) methodology for generating masks of parking spaces using a fleet of drones. However, a comprehensive evaluation of the accuracy of these masks has not been conducted. Furthermore, the implementation of this approach necessitates the acquisition of a top-view image of the parking lot, rendering it impractical for interior parking structures.

\subsubsection{2.2.2 Detection Approaches}  
In the realm of detection approaches, the model effectively carries out both detection and classification tasks within a unified process, utilizing a deep learning architecture, as opposed to segregating these jobs into two distinct processes. This approach ensures the preservation of flexibility and facilitates rapid inference. In this domain, the designing strategy involves regressessing a parking slot as a background or a regions of interest (ROI), and subsequently enhancing its classification score. The aforementioned technique can be categorized into two distinct types: two stage detectors and one stage detectors. 

For two stage detector, a detection model such as Faster RCNN \cite{ren2015faster} is used in initial stage to propose ROI and the status of parking spots are predicted in the subsequent stage. In the recent work \cite{kirtibhai2020faster}, the researchers exclusively employed a Faster RCNN model as the detector to identify a stationary vehicle within the parking lot and ascertain the rate of occupancy based on the predetermined capacity and location of a previously owned parking facility. Then another classification model is need to be implemented to predict the status of parking spots. This methodology mitigates the challenges by transforming them into the widely recognized task of car detection. The emergence of novel designs, such as YOLOv4 \cite{bochkovskiy2020yolov4} and RetinaNet \cite{lin2017focal}, has introduced new possibilities for effectively capturing small objects in a parking lot. The utilization of drone imagery is employed in \cite{hsieh2017drone}. The car identification in top-view perspective is conducted by employing the Faster RCNN and YOLO models, which are afterwards integrated with the layout proposal.

For one stage detector, the network integrates both tasks together. For example, in the work \cite{padmasiri2020automated}, a custom version of RetinaNet \cite{lin2017focal} is used as a one-stage detector. Nevertheless, the findings indicate a significant level of ambiguity in distinguishing between moving vehicles and parking spots that are currently occupied. To solve this problem, the authors in \cite{duong2022towards} proposed an attention mechanism on the parking lot region through grid-anchor regressions and use a custom versiom of MBN-FPN network \cite{sandler2018mobilenetv2} as the backbone. In the study \cite{ding2019vehicle}, the researchers suggest the incorporation of residual blocks into the Yolov3 architecture \cite{redmon2018yolov3} as a means to enhance the extraction of more detailed characteristics.

Our methodology adheres to the two-stage detector approach. In the subsequent phase, rather than employing a classification model to forecast the condition of a parking spot, we utilize the detector from the initial stage and implement it on frames that have undergone ROI-filtering (only showing the ROI). In the end, a straightforward mapping schema is executed in order to generate the status of all the parking slots simultaneously.

%% file: 3-Methodology.tex
\section{Methodology}
\label{sec:Methodology}

\begin{figure}[htb!]
\begin{center}
	\includegraphics[width=1\linewidth]{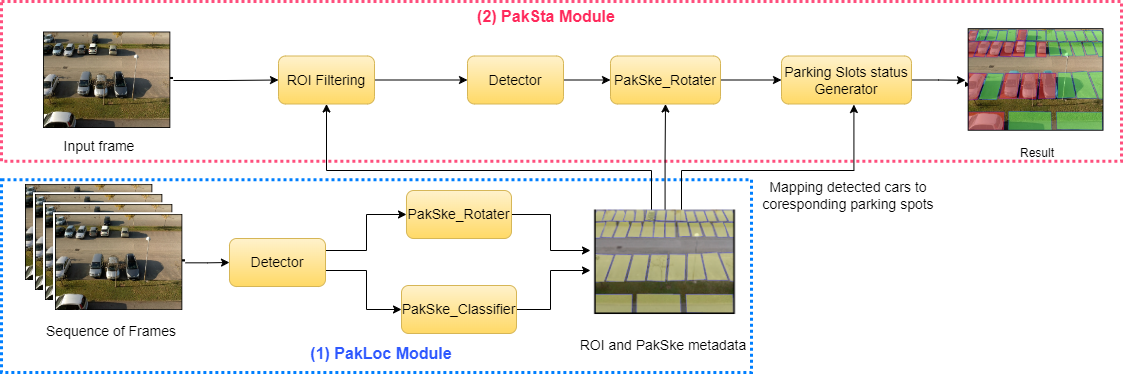}
	\caption{Proposed Architecture. It includes two main modules: \textcolor{blue}{(1) PakLoc} for automatic parking spots localization task and \textcolor{red}{(2) PakSta} for parking spots status identification task.}
	\label{fig:architecture}
\end{center}
\end{figure}

As describe in Section \ref{sec:introduction}, our proposed method is divided into two modules: (1) PakLoc for automatic parking spots localization task and (2) PakSta for parking spots status identification task. The detail architecture is visualized in Figure \ref{fig:architecture}. The automobile detector plays a vital role in our proposed concept. The component in question holds significant importance in both the PakLoc and PakSta modules, since its performance directly influences the overall consequences of the architecture. Consequently, we have partitioned this section into three distinct components, namely the Vehicle Detector, PakLoc, and PakSta.

\subsection{Vehicle Detector}
There exist two methodologies for deep learning detection frameworks: the CNN-based approach and the Transformer-based model. The efficacy of transformer-based models in extracting various diversified discriminative parts of information and fine-grained features has been demonstrated in \cite{he2021transreid}. Furthermore, the PKlot dataset \cite{de2015pklot} includes variations in car scale and viewpoint. As a result, in this study, the we employ a transformer-based object detection model known as deformable DETR \cite{zhu2020deformable} that demonstrates effective performance in detecting objects of varying scales. This detector model has been pre-trained using the VeRi-776 \cite{liu2016deep} and CityFlow \cite{tang2019cityflow} datasets. The VeRi-776 dataset has a total of 49,357 images depicting 776 distinct vehicles. These images were captured using 20 different cameras. On the other hand, the CityFlow dataset consists of over 229,680 labeled bounding boxes of 666 distinct cars. These bounding boxes were obtained from 40 cameras positioned at 10 different junctions. The detector was pre-trained for 70 epochs using the same set of parameters as described in the original research. The detector model is fine-tuned on the training set of the PKlot dataset for a total of 40 epochs.


\subsection{PakLoc - Automatic Parking Spots Localization}
\label{sec:method_pakloc}
We approach the problem of automatic parking spot localization as a problem of vehicle movement tracking over a number of consecutive frames. As depicted in Figure \ref{fig:architecture}, the PakLoc module utilizes successive frames as inputs to execute the car detection operation, resulting in the generation of bounding boxes for the discovered cars. Subsequently, the newly identified bounding boxes are compared to the existing inventory of currently tracked vehicles. The Intersection over Union (IoU) metric is employed to evaluate the relationship between a pair of bounding boxes. When two separate boxes do not intersect, their IoU value is 0. The IoU metric has a positive correlation with the extent of overlap between two bounding boxes. When a car remains stationary, the IoU value of the bounding boxes representing this car in consecutive frames will exhibit a high value. If this IoU value exceeds a predetermined IoU threshold denoted as $\theta$, a counter associated with that specific bounding box will be increased. If the counter above a predetermined frame threshold  $\gamma$, the vehicle will come to a halt for a specified duration, at which point the corresponding place will be designated as a parking spot. In the event that a bounding box derived from the present frame fails to correspond with any existing box within the tracked list, it can be inferred that a new vehicle has entered the scene. Consequently, this new vehicle is appended to the list of tracked cars. If a bounding box from the tracked list does not correspond to any newly discovered car after conducting a thorough comparison, it can be inferred that the bounding box is not stationary. Consequently, the bounding box in question will be eliminated from the list of objects being tracked. 

An effective boundary condition for distinguishing between a moving car and a parked car can be achieved by carefully selecting the IoU threshold $\theta$ and frame threshold $\gamma$. When a car is in motion, the IoU value between its current bounding box and the preceding one will gradually drop. Consequently, the car will eventually fail to meet the frame threshold. When the IoU threshold is set to a higher value, the algorithm's sensitivity to minor variations in the bounding box's position increases. This may lead to the identification of duplicate bounding boxes for a particular parking spot. Additionally, it is important to set the IoU threshold value at a sufficiently high level in order to disregard the junction of neighboring parking spots. Based on empirical findings, it has been determined that utilizing an IoU threshold $\theta$ within the range of $0.4$ to $0.9$ yields optimal outcomes. Our work undertakes an ablation investigation to determine the ideal IoU threshold $\theta$. The findings presented in Section \ref{sec:pakloc_result} indicate that the ideal value for $\theta$ is judged to be 0.75. Given that the PKLot dataset comprises photos that are recorded at a minimum interval of 5 minutes, a vehicle that remains stationary for a duration beyond 20 minutes would be considered classified as parked. Consequently, the value for $\gamma$ is set at 4. It is worth mentioning that the parameters $\theta$ and $\gamma$ need to be changed when applying our model to various datasets.

\begin{figure}[htb!]
\begin{center}
	\includegraphics[width=0.9\linewidth]{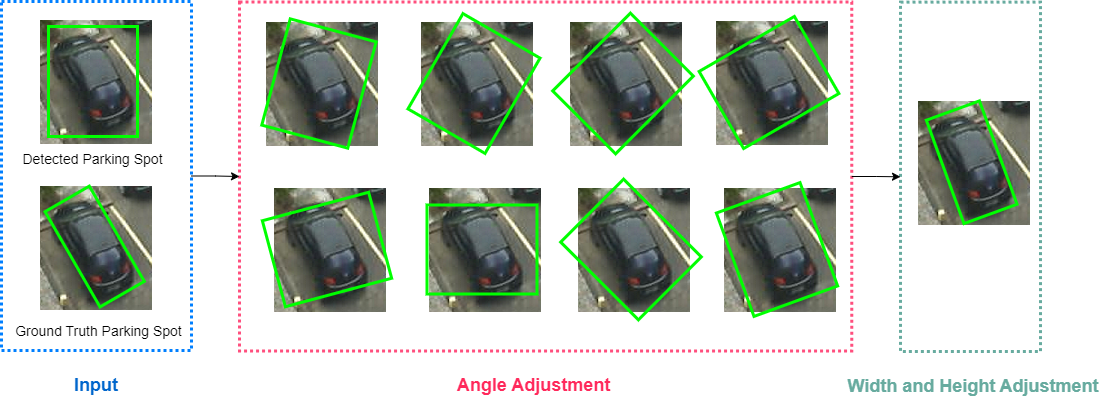}
	\caption{PakSke\_Rotater. Detected parking place and ground truth are input. By finding the greatest IoU value, it automatically adjusts parking slot angle, width, and height. }
	\label{fig:pakske}
\end{center}
\end{figure}

In the end, the PakSke layers are incorporated prior to generating the final outcome. The main goal of PakSke is to accurately align the detected parking spots with the actual angle of corresponding ground true parking spots, as parking spots often deviate from being perpendicular to the frame boundary (Figure \ref{fig:parking_spot}). The PakSke layers consist of two distinct branches: training branch and inference branch. Within the training branch, the PakSke\_Rotater layer is utilized to determine the optimal triplet hyperparameters, namely [angle, width\_scaling, height\_scaling] where $angle$ denotes the degree of rotation applied to the box and $width\_scaling, height\_scaling$ determine the proportional adjustment of the box's width and height, respectively. In other terms, a particular triplet propose and adjustment for a detected parking spot. The determination of the optimal triplet involves identifying the maximum IoU score between the generated parking spot and the matching ground truth label. The procedure is visually represented in Figure \ref{fig:pakske}. In this study, the parameter $angle$ is defined within the interval $[0,180]$ with an increment of 15 degrees, while the variables $width\_scaling$ and $height\_scaling$ are defined within the interval $[0.5, 1.5]$ with an increment of $0.1$. 

While the PakSke\_Rotater layer may readily generate modified parking places, it is not feasible to employ this layer during the inference stage or deployment on fresh datasets due to the absence of ground truth labels. In order to address this issue, we suggest employing the PakSke\_classifier during the inference stage. The PakSke\_classifier is a simple CNN classifier that receives the bounding box picture of the identified parking spot as input and predicts the triplet $[angle, width\_scaling, height\_scaling]$. The PakSke\_classifier is learned during the training phase using the output produced by the PakSke\_Rotater layer. In this research endeavor, the implementation of PakSke\_classifier is executed utilizing the CmAlexnet architecture, as proposed in \cite{rahman2020convolutional} . A minor modification is made to the last layer of the network to generate a triplet [angle, width\_scaling, height\_scaling].

The ultimate outcome of the PakLoc module is the generation of Parking Spots metadata, which consists of a list of [CameraId,$ $x,y,w,h,$ $angle, width\_scaling, height\_scaling]. The $CameraId$ represents the unique identifier assigned to each camera. The notation $[x,y,w,h]$ denotes the standard coordinate system used to describe the position and dimensions of a parking spot. Similarly, the triplet $[angle, width\_scaling, height\_scaling]$ represents the set of PakSke parameters corresponding to the current parking spot.

\subsection{PakSta - Parking Spots Status Identification}
The objective of PakSta is to forecast the state of parking spaces identified from the PakLoc. The PakSta system receives camera video input in a sequential manner, processing each frame individually. These frames are then sent via a region of interest (ROI) filtering layer. The ROI in question corresponds to the coordinates of parking spots, which are obtained from the parking spot metadata extracted in PakLoc. Subsequently, the filtered image is inputted into the detector in order to identify the presence of a car within the image. In the parking slots status generator layer, every identified vehicle is assigned a parking spot in the parking spot metadata and its bounding box is adjusted using the appropriate triplet parameter $[angle, width\_scaling, height\_scaling]$. Ultimately, the parking spaces that are mapped to any identified vehicle are categorized as occupied, whereas the remaining spaces are categorized as vacant.

%% file: 4-Experiment_and_Result.tex
\section{Experiments and Results}
\label{sec:experiment}

\subsection{Dataset }
\begin{figure}[htbp]
\begin{center}
	\includegraphics[width=1\linewidth]{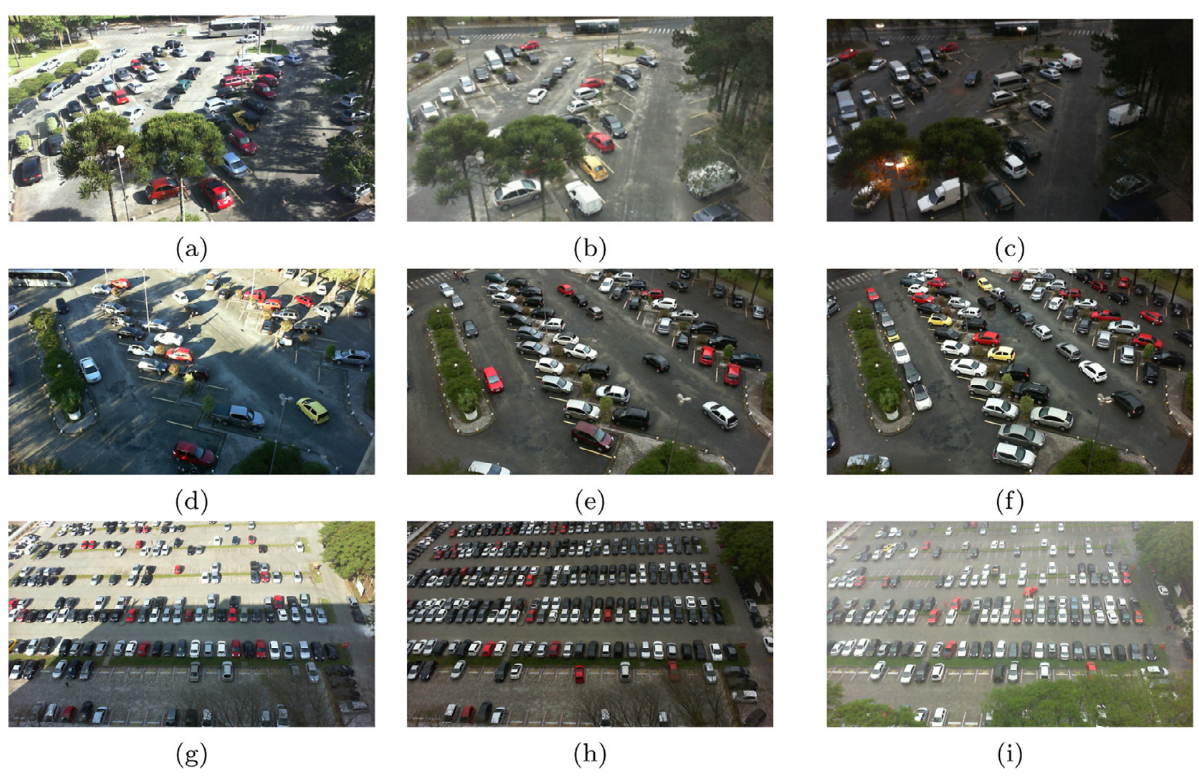}
	\caption{ Parking lot images captured under different weather and light conditions from Pklot Dataset \cite{de2015pklot} - (a) sunny (b) overcast, and (c) rainy from UFPR04; (d) sunny (e) overcast, and (f) rainy from UFPR05; and (g) sunny (h) overcast, and (i) rainy from PUCPR}
	\label{fig:pklot_all}
\end{center}
\end{figure}

\textbf{PKlot Dataset \cite{de2015pklot}}: PKlot is our primary dataset for method testing. The PKlot Dataset consists of 12,417 images captured from three angles in two parking lots. UFPR04 and UFPR05 subsets were captured by a camera on the fourth and fifth floors of the UFPR building.The cameras on the 10th floor of the Pontifical Catholic University of Parana acquired the third parking lot subset, specifically (PUCPR). 
The dataset consists of 695,851 manually identified parking spaces, of which 337,780 are occupied (48.6\%) and 358,071 are vacant (51.4\%). The standard image dimension is 1280 720 pixels. Each image is accompanied by an XML file containing four points representing the polygons of the monitored parking spots. The XML files also display the occupancy status of each parking space.
The majority of UFPR04 and UFPR05 parking locations are labeled. On the contrary, the PUCPR manually identified 100 parking spaces, while the images estimate 300 parking spaces perceptible to the human eye. Each image was classified as sunny, rainy, or cloudy and was captured at 5-minute intervals during daytime hours. Figure \ref{fig:pklot_all} displays images of parking lots and weather conditions from the PKLot dataset. 
In compliance with the original publication \cite{de2015pklot}, in this study, the PKLot dataset was divided into a training and testing set for this investigation . Details are available in Table \ref{tab:test_data}.

\begin{table}[htbp]
\centering
\begin{tabular}{ll|lll|lll}
\textbf{}                             &                   & \multicolumn{3}{c|}{\textbf{Training sets}}                                                   & \multicolumn{3}{c}{\textbf{Testing sets}}                                                    \\ \cline{3-8} 
                                      & \textbf{}         & \multicolumn{1}{l|}{\textbf{Occupied}} & \multicolumn{1}{l|}{\textbf{Empty}} & \textbf{Total} & \multicolumn{1}{l|}{\textbf{Occupied}} & \multicolumn{1}{l|}{\textbf{Empty}} & \textbf{Total} \\ \hline
\multicolumn{1}{l|}{\textbf{UFPRO4}}  & \textbf{Sunny}    & \multicolumn{1}{l|}{16,524}            & \multicolumn{1}{l|}{14,327}         & 30,851         & \multicolumn{1}{l|}{15,642}            & \multicolumn{1}{l|}{12,007}         & 27,649         \\ \cline{2-8} 
\multicolumn{1}{l|}{}                 & \textbf{Overcast} & \multicolumn{1}{l|}{6989}              & \multicolumn{1}{l|}{15,076}         & 22,065         & \multicolumn{1}{l|}{4619}              & \multicolumn{1}{l|}{12,703}         & 17,322         \\ \cline{2-8} 
\multicolumn{1}{l|}{}                 & \textbf{Rainy}    & \multicolumn{1}{l|}{1041}              & \multicolumn{1}{l|}{2553}           & 3594           & \multicolumn{1}{l|}{1310}              & \multicolumn{1}{l|}{3054}           & 4364           \\ \cline{2-8} 
\multicolumn{1}{l|}{}                 & \textbf{Total}    & \multicolumn{1}{l|}{24,554}            & \multicolumn{1}{l|}{31,956}         & 56,510         & \multicolumn{1}{l|}{21,571}            & \multicolumn{1}{l|}{27,764}         & 49,335         \\ \hline
\multicolumn{1}{l|}{\textbf{UFPROS,}} & \textbf{Sunny}    & \multicolumn{1}{l|}{28,822}            & \multicolumn{1}{l|}{21,657}         & 50,479         & \multicolumn{1}{l|}{28,762}            & \multicolumn{1}{l|}{20,649}         & 49,411         \\ \cline{2-8} 
\multicolumn{1}{l|}{}                 & \textbf{Overcast} & \multicolumn{1}{l|}{15,421}            & \multicolumn{1}{l|}{12,985}         & 28,406         & \multicolumn{1}{l|}{18,343}            & \multicolumn{1}{l|}{10,217}         & 28,560         \\ \cline{2-8} 
\multicolumn{1}{l|}{}                 & \textbf{Rainy}    & \multicolumn{1}{l|}{2751}              & \multicolumn{1}{l|}{1633}           & 4384           & \multicolumn{1}{l|}{3327}              & \multicolumn{1}{l|}{1218}           & 4545           \\ \cline{2-8} 
\multicolumn{1}{l|}{}                 & \textbf{Total}    & \multicolumn{1}{l|}{46,994}            & \multicolumn{1}{l|}{36,275}         & 83,269         & \multicolumn{1}{l|}{50,432}            & \multicolumn{1}{l|}{32,084}         & 82,516         \\ \hline
\multicolumn{1}{l|}{\textbf{PUCPR}}   & \textbf{Sunny}    & \multicolumn{1}{l|}{47,490}            & \multicolumn{1}{l|}{59,731}         & 107,221        & \multicolumn{1}{l|}{49,271}            & \multicolumn{1}{l|}{51,941}         & 101,212        \\ \cline{2-8} 
\multicolumn{1}{l|}{}                 & \textbf{Overcast} & \multicolumn{1}{l|}{26,774}            & \multicolumn{1}{l|}{42,933}         & 69,707         & \multicolumn{1}{l|}{15,589}            & \multicolumn{1}{l|}{47,484}         & 63,073         \\ \cline{2-8} 
\multicolumn{1}{l|}{}                 & \textbf{Rainy}    & \multicolumn{1}{l|}{19,540}            & \multicolumn{1}{l|}{16,025}         & 35,565         & \multicolumn{1}{l|}{35,565}            & \multicolumn{1}{l|}{11,926}         & 47,491         \\ \cline{2-8} 
\multicolumn{1}{l|}{}                 & \textbf{Total}    & \multicolumn{1}{l|}{93,804}            & \multicolumn{1}{l|}{118,689}        & 212,493        & \multicolumn{1}{l|}{100,425}           & \multicolumn{1}{l|}{111,351}        & 211,776        \\ \hline
\textbf{Total of samples}                      &                   & \multicolumn{1}{l|}{165,352}           & \multicolumn{1}{l|}{186,920}        & 352,272        & \multicolumn{1}{l|}{172,428}           & \multicolumn{1}{l|}{171,199}        & 343,627       
\end{tabular}
\caption{Training and Testing sets in PKLot dataset}
\label{tab:test_data}
\end{table}

\textbf{CityFlow Dataset \cite{tang2019cityflow}}: The detector is pre-trained using CityFlow \cite{tang2019cityflow}. CityFlow contains 229,680 labeled bounding boxes for 666 automobiles from 40 cameras at 10 intersections. 3.25 hours of traffic were logged for the training and validation datasets. Test data contain twenty minutes of traffic video. 

\textbf{Veri-776 Dataset \cite{liu2016deep}}: Similar to CityFlow, the Veri-776 dataset is utilized for the detector's initial training. VeRi-776 contains 49,357 images of 776 automobiles. The photos are obtained from unrestricted real-world traffic scenarios and tagged with bounding boxes, categories, colors, and brands. Each vehicle is observed by two to eighteen cameras. These cameras are strategically situated to capture vehicles from various angles, lighting conditions, and occlusions.

\subsection{Evaluation Metrics}
\label{sec:evaluation_metric}
For the automatic parking spots localization problem (PakLoc), the objective is to accurately identify the maximum number of parking spots. As the result, the most suitable metric for evaluating the performance of the detection system is recall. In this study, we present the average recall at an IoU threshold of 0.75, denoted as AR75. Additionally, we report the mean average recall for IoU thresholds ranging from 0.4 to 0.9, with a step size of 0.05, referred to as mAR40\_90. Furthermore, for the purpose of comparing the performance of our proposed method to that of related work, we present the average precision at an IoU threshold of 0.5 (AP50).

Similarly, as we wish to maximize the accuracy of the model when identifying the status of parking spaces, precision becomes the appropriate metric. Specifically, we select the metric as average precision at the IoU threshold $\theta$ of 0.75 (AP75).
 


\subsection{Baselines}
\label{sec:baselines}
As discussed in Section \ref{sec:related_work}, there is no default benchmark setting for the problem of automatic parking spots localization. The existing work shows their performance in different datasets with different metrics. We then select some typical work \cite{kirtibhai2020faster, patel2020car, padmasiri2020automated, de2023vehicle} that show their performance with a appropriate metric for the evaluation. 

In the context of identifying the status of parking spots, the performance of PakSta is compared to that of other relevant studies\cite{duong2022towards, padmasiri2020automated, redmon2018yolov3}. The specifics of these baseline models were discussed in Section \ref{sec:related_work}.


\subsection{Experimental Results}
\label{sec:result}
This section is divided into two subsection to evaluate the performance of PakLoc and PakSta.

\subsubsection{PakLoc Performance}
\label{sec:pakloc_result}
The evaluation of PakLoc is conducted using the test set of the PKLot dataset. Firstly, as discussed in Section \ref{sec:method_pakloc}, an ablation study is conducted to determine the best parameter IoU threshold $\theta$. In this ablation study, we assess the results (AR) of PakLoc by varying the parameter $\theta$ throughout the range of 0.4 to 0.9, with a step size of 0.05. The outcome presented in Table \ref{tab:pakloc_theta} indicates that the ideal value of $\theta$ is determined to be 0.75. Then, we select the $\theta$ as $0.75$ for all next experiment. It is worth noting that in this ablation investigation, we adjusted the frame threshold $\gamma$ to 4.

\begin{table}[htbp]
\centering
\begin{tabular}{l|l}
\multicolumn{1}{c|}{\textbf{IoU Threshold $\theta$}} & \multicolumn{1}{c}{\textbf{AR}} \\ \hline
0.4                                           & 59.88                           \\ \hline
0.45                                          & 61.77                           \\ \hline
0.5                                           & 70.55                           \\ \hline
0.55                                          & 77.88                           \\ \hline
0.6                                           & 84.14                           \\ \hline
0.65                                          & 89.05                           \\ \hline
0.7                                           & 92.53                           \\ \hline
0.75                                          & \textbf{94.25}                           \\ \hline
0.8                                           & 93.98                           \\ \hline
0.85                                          & 91.72                           \\ \hline
0.9                                           & 87.46                          
\end{tabular}
\caption{PakLoc result on testset of PKLot with different IoU threshold $\theta$}
\label{tab:pakloc_theta}
\end{table}

As mentioned in Section \ref{sec:evaluation_metric}, three metrics are employed in this study, namely AR75, mAR40\_90, and AP50. Then, to demonstrate the impact of the PakSke layer, we present outcomes obtained from both scenarios: one with the inclusion of the PakSke layer and the other without it using these three metrics. The results in Table \ref{tab:pakske_pakloc} demonstrates the efficiency of our proposed PakSke layers.  Using PakSke layers increases all three metrics by at least 6\%.

\begin{table}[htb!]
\centering
\begin{tabular}{l|l|l|l}
                        & \textbf{AR75} & \textbf{mAR40\_90} & \textbf{AP50} \\ \hline
\textbf{Without PakSke} & 88.31         & 74.38              & 80.23         \\ \hline
\textbf{With PakSke}    & \textbf{94.25}         & \textbf{82.11}              & \textbf{86.37}        
\end{tabular}
\caption{PakLoc result on testset of PKLot with and without PakSke layer}
\label{tab:pakske_pakloc}
\end{table}

Lastly, we compare PakLoc's performance to other baselines described in Section \ref{sec:baselines}. The results presented in Table \ref{tab:pakloc_final} indicate that our proposed method outperforms all prior work using the same dataset with 86.4\% AP50. It even achieve a better result with 92.7\% AP75.

\begin{table}[htb!]
\centering
\begin{tabular}{l|l|l|l|l}
\textbf{Method/Ref}  & \textbf{Backbone}        & \textbf{Test Set} & \textbf{Metric} & \textbf{Result} \\ \hline
Faster PSP \cite{kirtibhai2020faster}            & Faster-RCNN              & CNRPark-EXT   & AP50            & 83.1            \\ \hline
Auto PSP \cite{patel2020car}            & Yolo4                    & CNRPark-EXT   & AP50            & \textbf{97.6}   \\ \hline
Realtime PSP \cite{padmasiri2020automated}            & Resnet   and faster RCNN & PKLot         & AP50            & 63.6            \\ \hline
Cascade PSP \cite{de2023vehicle}            & Cascade Mask R-CNN       & PKLot         & AP50            & 59.1            \\ \hline
PakLoc (ours) & Deformable DETR          & PKLot         & AP50            & 
\textbf{86.4} \\ \hline
PakLoc (ours) & Deformable DETR          & PKLot         & AP75            & 
\textbf{92.7}  

\end{tabular}
\caption{PakLoc result on testset of PKLot and other baselines method}
\label{tab:pakloc_final}
\end{table}

\subsubsection{PakSta Performance}
As outlined in Section \ref{sec:baselines}, a comparison is made between the results of PakStat and three additional baseline models \cite{duong2022towards, padmasiri2020automated, redmon2018yolov3}. The findings are presented in table \ref{tab:paksta_final}. The approach presented in \cite{duong2022towards} achieved the highest AP75 score of 98\%. However, it should be noted that this system relied on the manual annotation of parking places during the training phase. This factor restricts the practical implementation of the paradigm in real-world scenarios. In contrast, our suggested solution, PakSta, does not necessitate human labeling of parking slots for implementation in a fresh dataset or a real parking lot environment. Furthermore, PakSta was able to obtain a notable outcome of 93.6\% AP75, positioning it as the second best performer. This even surpasses the strategy employed in the study \cite{redmon2018yolov3}, where manually labeled parking spaces data was utilized. In addition, the results table further demonstrates the effectiveness of PakSke layers by indicating that they contribute a positive impact (improve 6\%) on the ultimate outcome of PakSta.

\begin{table}[htb!]
\centering
\begin{tabular}{l|l|l|l|c}
\multicolumn{1}{c|}{\textbf{Method/Ref}} & \multicolumn{1}{c|}{\textbf{Backbone}} & \multicolumn{1}{c|}{\textbf{Data}} & \multicolumn{1}{c|}{\textbf{\begin{tabular}[c]{@{}c@{}}Use Manually Label \\ Of Parking Spots\end{tabular}}} & \textbf{AP75} \\ \hline
POD \cite{padmasiri2020automated}                                       & RetinaNet                              & PKLot                              & No                                                                                                           & 61.8         \\ \hline
Yolo SPS \cite{redmon2018yolov3} & Yolo3                                  & PKLot                              & Yes                                                                                                          & 93.3          \\ \hline
OcpDept \cite{duong2022towards}                               & MBN-FPN       & PKLot                              & Yes                                                                                                          & \textbf{98.0}            \\ \hline
PakSta without PakSke  (ours)                   & Deformable DETR                        & PKLot                              & No                                                                                                           & 87.3         \\ \hline
PakSta with PakSke (ours)                         & Deformable DETR                        & PKLot                              & No                                                                                                           & \textbf{93.6}       
\end{tabular}
\caption{PakSta result on testset of PKLot and other baselines method}
\label{tab:paksta_final}
\end{table}

%% file: 5-Conclusion.tex
\section{Conclusion}
\label{sec:conclusion}
In this paper, two new approaches, PakLoc and PakSta, are proposed to address the problems of automatic parking spot localization and parking spot status identification, respectively. Both of these methods demonstrate superior performance compared to the existing approaches in the given context. Additionally, we propose the incorporation of PakSke layers as a means to enhance the performance of these methods. The utilization of PakSke layers as a plug-in module is applicable in a comparable manner. In the forthcoming period, our objective is to construct a comprehensive smart parking system utilizing the way we have put forth.

%% file: 6-Acknowledgement.tex
\section*{Acknowledgement}

This material is based upon work partially supported by the U.S. Department of Energy’s Office of Energy Efficiency and Renewable Energy (EERE) under the Award Number DE-EE0009208. The views and opinions of authors expressed herein do not necessarily state or reflect those of the United States Government or any agency thereof.


